\documentclass{article}
\usepackage{spconf,amsmath,graphicx}
\usepackage{color}

\definecolor{Red}{rgb}{1, 0.2, 0.2}

\title{Deep LOGISMOS: Deep Learning Graph-based 3D Segmentation of Pancreatic Tumors on CT scans}
\name{Zhihui~Guo$^{1}$, Ling Zhang${^{2*}}$, Le Lu${^{2}}$, Mohammadhadi Bagheri$^{2}$, Ronald M. Summers$^{2}$,  \thanks{*ling.zhang3@nih.gov; milan-sonka@uiowa.edu. This research was supported -- in part -- by  Intramural Research Program of  National Institutes of Health Clinical Center and NIH R01 EB004640. The authors thank Nvidia for the TITAN X Pascal GPU donation.}}
\address{$^{1}$Iowa Institute for Biomedical Imaging, The University of Iowa, Iowa City IA \\
	 $^{2}$Radiology and Imaging Sciences Department, National Institutes of Health (NIH), Bethesda MD}
\begin{document}
%\ninept
%
\maketitle
\begin{abstract}
This paper reports Deep LOGISMOS approach to 3D tumor segmentation  by incorporating boundary information derived from deep contextual learning to LOGISMOS -- layered optimal graph image segmentation of multiple objects and surfaces. Accurate and reliable tumor segmentation is essential to tumor growth analysis and treatment selection. 
%In this paper, we propose a 3D tumor segmentation framework that incorporates deep learning into a layered optimal graph image segmentation of multiple objects and surfaces (called Deep LOGISMOS) . 
A fully convolutional network (FCN), UNet, is first trained using three adjacent 2D patches centered at the tumor, providing contextual UNet segmentation and probability map for each 2D patch. The UNet segmentation is then refined by Gaussian Mixture Model (GMM) and morphological operations. The refined UNet segmentation is used to provide the initial shape boundary to build a segmentation graph. The cost for each node of the graph is determined by the UNet probability maps. Finally, a max-flow algorithm is employed to find the globally optimal solution thus obtaining the final segmentation. For evaluation, we applied the method to pancreatic tumor segmentation on a dataset of 51 CT scans, among which 30 scans were used for training and 21 for testing. With Deep LOGISMOS, DICE Similarity Coefficient (DSC) and Relative Volume Difference (RVD) reached 83.2$\pm$7.8\% and 18.6$\pm$17.4\% respectively, both are significantly improved (p$<$0.05) compared with contextual UNet and/or LOGISMOS alone.
\end{abstract}
\section{Introduction}
\label{sec:intro}
Monitoring the growth and spread of tumors at different time points helps physicians differentiate tumor types and plan the proper treatment \cite{lingMiccai}. To achieve this, accurate and reliable segmentation of tumors is of great importance.

%------------------------------------------------ Figure ----------------------------------
\begin{figure*}[hpt]
\includegraphics[width=17.8cm]{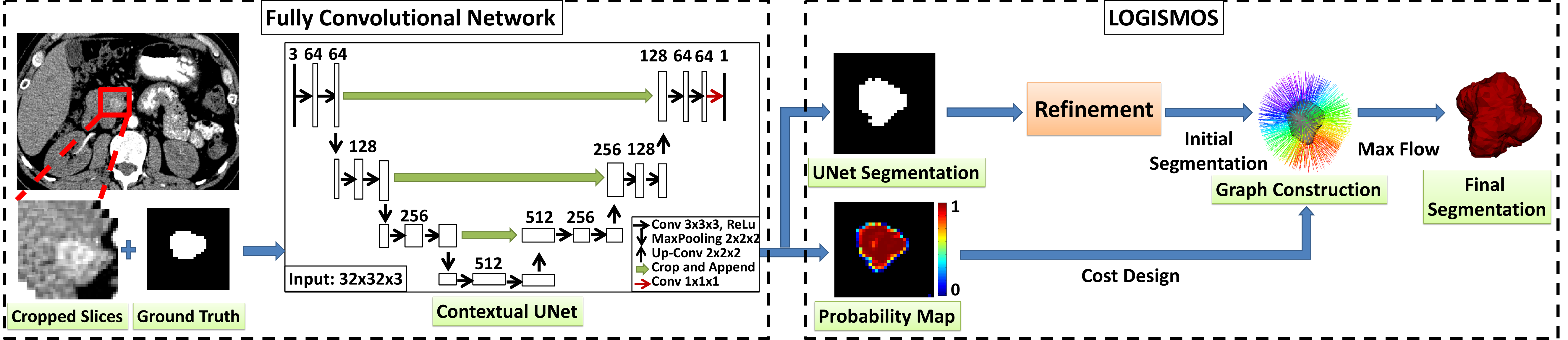}
\caption[example] 
{ \label{flowchart} 
Schematic diagram of Deep LOGISMOS method illustrated on an example of 3D pancreatic tumor segmentation. }
\end{figure*} 
%-------------------------------------------------------------------------------------------

Variety of methods have been proposed for medical image segmentation, among which deep learning has recently become prevalent  and reached new levels of the state-of-the-art accuracy in many tasks \cite{rnn2016NIPS}. Medical imaging data such as CT and MRI are inherently 3D, but can be visualized as stacks of 2D slices. Deep learning based segmentation methods can be divided into 4 categories according to how data is input to the network: convolutional neural networks (CNN) with 2D convolutions \cite{Unet}; CNN with 3D convolutions \cite{3dUnet}; combination of 2D CNN and recurrent neural networks (RNN) for 3D segmentation \cite{rnn2016NIPS, fcnRNN2017CVPR}; and combination of 2D CNN and optimization algorithm for 2D or 3D segmentation \cite{ling2017isbi}. UNet \cite{Unet} is a type of CNN with 2D convolutions that only takes intra-slice context into account, leaving out the inter-slice context. 3D UNets \cite{3dUnet} apply 3D convolutions to capture 3D spatial context but are computationally expensive. An alternative way to use contextual UNet is to stack three adjacent slices in three RGB channels to leverage inter-slice context from adjacent slices. LSTM \cite{lstm} is a type of RNN that is designed for sequential data and can be used to leverage spatial context between adjacent slices. Chen \textit{et al}. \cite{rnn2016NIPS} combined a modified 2D UNet and LSTM to do 3D segmentation of neuron and fungus. Tseng \textit{et al.} \cite{fcnRNN2017CVPR} applied CNN and convolutional LSTM to multi-modality data and achieved 3D segmentation. In \cite{ling2017isbi}, FCN and graph-based method worked together where FCN provided the cost for the 2D graph. Since objects are 3D in nature, 3D spatial context are valuable in 3D segmentation.  

LOGISMOS is a graph-based framework that translates geometric constraints of interacting surfaces and objects into graph arcs and likelihood of segmentation surface positioning into graph node/arc costs \cite{yinLogismos}. With LOGISMOS, the globally optimal N-dimensional solution satisfying defined smoothness constraints is obtained. The node-weighted LOGISMOS has been successfully used in difficult tasks such as 3D knee and brain segmentation \cite{yinLogismos, guoLogismos}. This graph segmentation framework is robust to image noise and weak boundaries but requires a proper initial segmentation as the shape prior to build the graph and to assign proper costs for each node reflecting its likelihood to occur on the desired segmentation surface. Defining the initial segmentation is not trivial and often requires manual intervention. Similarly, the graph costs are frequently derived from hand-crafted  task-specific features and may not be generalizable to other problems.

We propose a method that combines UNet and LOGISMOS for 3D tumor segmentation. Specifically, we adopt UNet to integrate intra-slice and adjacent-slice contexts, and regulate the 3D shape by LOGISMOS. Different from our previous FCN+graph method \cite{ling2017isbi} which uses FCN to locate the object center for graph construction and combines FCN-derived cost with hand-crafted costs, our method directly constructs the graph based on the UNet-derived object boundaries and assigns UNet-derived probabilities as costs.

Pancreatic cancer is a major health problem that shows a steady increase in incidence and death rate while also exhibiting a slight improvement in survival rates over the past 5 years \cite{siegel2014cancer}. 
To our best knowledge, this is the first approach for automated 3D segmentation of pancreatic tumors. The proposed method can be extended to any tumor segmentation tasks.

\section{Methods}
\label{sec:methods}
We present a method called Deep LOGISMOS to segment tumors in 3D by combining contextual UNet and a graph-based framework LOGISMOS. The work-flow is described in Fig.\ref{flowchart}. First, the tumor ROI, defined as a square cube (32$\times$32$\times$32 voxels) by a single click of its center point, is cropped from the whole image. For each 2D slice, the contextual UNet takes itself and its two adjacent 2D slices as input patch and outputs the probability map and segmentation. We apply a GMM to remove false positives. After that, morphological opening and closing are applied to retain only the largest region in the segmentation. To construct the graph, the refined UNet segmentation is set as the initial segmentation to build the graph. The UNet probability maps is used as the cost for nodes in the graph. The final segmentation is given by the global optimal solution via a max-flow algorithm in graph search \cite{boykov2004}.  

\subsection{Contextual UNet}
With the multi-scale training architecture, UNet meets the need for biomedical image segmentation and has achieved great success in various tasks \cite{rnn2016NIPS}. We use the UNet described in \cite{Unet} for end-to-end training in this study, with the modification that the lowest scale of feature maps is removed due to small 2D image size (32 $\times$ 32). The input is 3-adjacent 2D slices, leveraging the adjacent spatial contexts, namely contextual UNet. To increase training sample size, data augmentations including translation, rotation and scaling  are applied to each sample. The initial learning rate is set as 1e-6 with momentum optimizer. We train UNet for around 30 epochs. We test several batch size options (1, 3, 10, 100) in the verification, and the batch size of 1 gives the best accuracy. 

\subsection{Refinement}
\label{sub:refine}
The UNet output needs to be further refined due to two reasons. First, the intensity distributions of tumors vary greatly for different patients and different contrast phase. Since the training set is small, the diverse intensity distributions may compromise the performance of UNet. Second, the purpose is to segment the center tumor inside the ROI. However, there may be other tumors in the image that are detected by UNet and should be excluded. We adopt a GMM with prior information about the relative intensity distributions of tumors and background to subtract background from UNet segmentation. Afterwards, morphological opening and closing are applied to ensure that only the largest region in the center is retained.
%\subsubsection{False Positive Reduction}
%\label{fpreduction}

For false positive reduction, only pixels that are segmented as tumors by UNet inside the ROI are considered. We fit two Gaussian distributions with GMM from all pixel intensities. GMM is a clustering method that applies maximum likelihood estimation with Gaussian conditional distribution and is solved by Expectation-Maximization algorithm. The motivation to fit two Gaussian distributions (N($\mu_{1}$, $\sigma_{1}^2$), N(($\mu_{2}$, $\sigma_{2}^2$)) for tumor and background respectively is based on the prior information that pixels inside one tumor have relatively homogeneous intensities, which are higher than intensities in the background. Suppose $\mu_{1}$ is larger than $\mu_{2}$, the condition to apply the false positive reduction by GMM is $\mu_{2}<\mu_{1}-\sigma_{1}$. If the condition is satisfied, pixels with intensities less than $\mu_{2}$ are marked as background and the probabilities are set to be 0. Otherwise, no false positive reduction will be applied.
%\begin{equation}
%\label{eq:FPreduction}
%\mu_{2}<\mu_{1}-\sigma_{1}
%\end{equation}
Then, two iterations of 3D morphological opening are applied and only the largest region is kept. Afterwards, 3D morphological closing is performed.

%\subsubsection{Morphological process}
%To retain only the center tumor inside the ROI, 2 iterations of 3D morphological opening are first applied. Through opening, the bridged tumors are separated into non-connected regions. We only keep the largest region near the ROI center. Afterwards, 3D morphological closing is used to fill holes and return the region to the original size. In this way, small separated regions are removed and only the center tumor remains in the ROI.

\subsection{LOGISMOS}
There are two key factors that affect the performance of our graph-based method, namely the initial segmentation and the cost design. We take advantage of UNet to generate a reliable initial segmentation and assign costs from deep features.
\subsubsection{Graph construction}
The UNet segmentation after refinement can be regarded as a coarse initial segmentation. This type of initial segmentation contains image-specific shape information of the tumor on an unseen image, which is preferable to be the shape prior compared with simple shape such ellipse or a mean shape model. Based on the boundary of initial segmentation, a geometric node-weighted graph is established. A stack of graph nodes (called a column) are connected with intra-column arcs that ensure only one cut through the column. Besides, inter-column arcs encode the smoothness constraints. The columns are built starting from the normal directions of points on the boundary under electric lines of force (ELF) \cite{yinLogismos} to avoid intersection. The length of the columns is set as 50 with node spacing of 0.5 mm to cover the potential area of the tumor.
%%------------------------------------------------ Figure ----------------------------------
%\begin{figure}[hpt]
%\includegraphics[width=8cm]{figs/graph.pdf}
%\centering
%\caption[example] 
%{ \label{slices} 
%Schematic of graph construction. (a). Graph overlaid with initial segmentation and three image planes. The color indicates column id. (b). Simplified illustration of nodes and arcs for three adjacent columns. The base arcs determine the graph domain.}
%\end{figure} 
%-------------------------------------------------------------------------------------------
\subsubsection{Cost design}
Contextual UNet outputs a probability map for each 2D slice. The probability is a region-based likelihood that ranges from 0 to 1 with higher value indicating higher chance of the pixel to be inside the tumor. LOGISMOS requires the cost to be the likelihood of nodes being not on the boundary. To translate the region-based probabilities to boundary-based cost, Eq.\ref{eq:cost} is used to decide the cost for node j on column k based on the summation of the probabilities for interior nodes on the same column. The -0.5 term corresponds to the probability threshold (0.5) when generating UNet segmentation. 
\begin{equation}
\label{eq:cost}
c_{n_{k,j}}=-\sum_{i=1}^j(p(n_{k,i})-0.5)
\end{equation}
\subsubsection{Segmentation}
The constructed graph integrates shape prior, geometric smoothness constrains, deep features, and ensures globally optimized true 3D segmentation. The final segmentation is obtained by max-flow algorithm \cite{boykov2004} in polynomial time.

\section{Experimental Results}
\label{sec:result}
Deep LOGISMOS was applied to a dataset of 51 arterial phase CT scans from 15 patients with pancreatic tumors studied at multiple time points, patients were participating in a clinical trial. The CT scans have a resolution of 1$\times$1$\times$1 $mm^3$ after resampling. A pancreatic tumor ROI with the size of 32$\times$32$\times$32 voxels was extracted from a scan by a single click at the approximate tumor center. 30 tumor ROIs from 8 patients were used for training and 21 tumor ROIs from other 7 patients were used for testing. We assessed the effect of 3 main aspects, which were the adjacent-slice context, refinement, and true 3D constraints of LOGISMOS. The evaluation metrics include DSC and RVD (relative volume difference). Statistical significance was estimated using paired t-test and the significance level was set at 0.05. The UNet is implemented using the Caffe platform \cite{jia2013caffe} on a Nvidia TITAN X Pascal GPU with 12 GB of memory.

\subsection{Contextual UNet vs.\ 2D UNet}
Besides training a contextual UNet, we also trained a 2D UNet on the same training set. All the parameters were the same. The performance of the two networks on the test set are presented in Table \ref{unet}. Contextual UNet achieved significantly superior segmentation accuracy than 2D UNet for RVD.

%------------------------------------------ Table -----------------------------------
\begin{table}[t]
\centering
\caption{Comparison of 2D UNet and contextual UNet.}
\label{unet}
\begin{tabular}{|c|c|c|}
\hline
Methods         & DSC (\%)       & RVD (\%)        \\ \hline
2D UNet         & 72.8 $\pm$ 22.0 & 42.5 $\pm$ 32.8 \\  \hline
\textbf{Contextual UNet} & 75.6 $\pm$ 16.6  & \textbf{35.7 $\pm$ 29.7}  \\   \hline
\end{tabular}
\end{table}
%-----------------------------------------------------------------------------------------

\subsection{Refinement}
Next, we compared the refined UNet segmentation with the original UNet segmentation (Table \ref{tb:refinement}). DSC and RVD indices demonstrated that segmentation with refinement was significantly better than that without refinement.
%------------------------------------------ Table -----------------------------------
\begin{table}[t]
\centering
\caption{Performance of contextual UNet segmentation.}
\label{tb:refinement}
\begin{tabular}{|c|c|c|}
\hline
Methods         & DSC (\%)       & RVD (\%)        \\ \hline
Without refinement         & 75.6 $\pm$ 16.6 & 35.7 $\pm$ 29.7 \\  \hline
\textbf{With refinement} & \textbf{81.6 $\pm$ 11.2}  & \textbf{26.1 $\pm$ 20.9}  \\   \hline
\end{tabular}
\end{table}
%-----------------------------------------------------------------------------------------

\subsection{Deep LOGISMOS vs.\ Contextual UNet, LOGISMOS}
\begin{figure}[t]
\includegraphics[width=7.5cm]{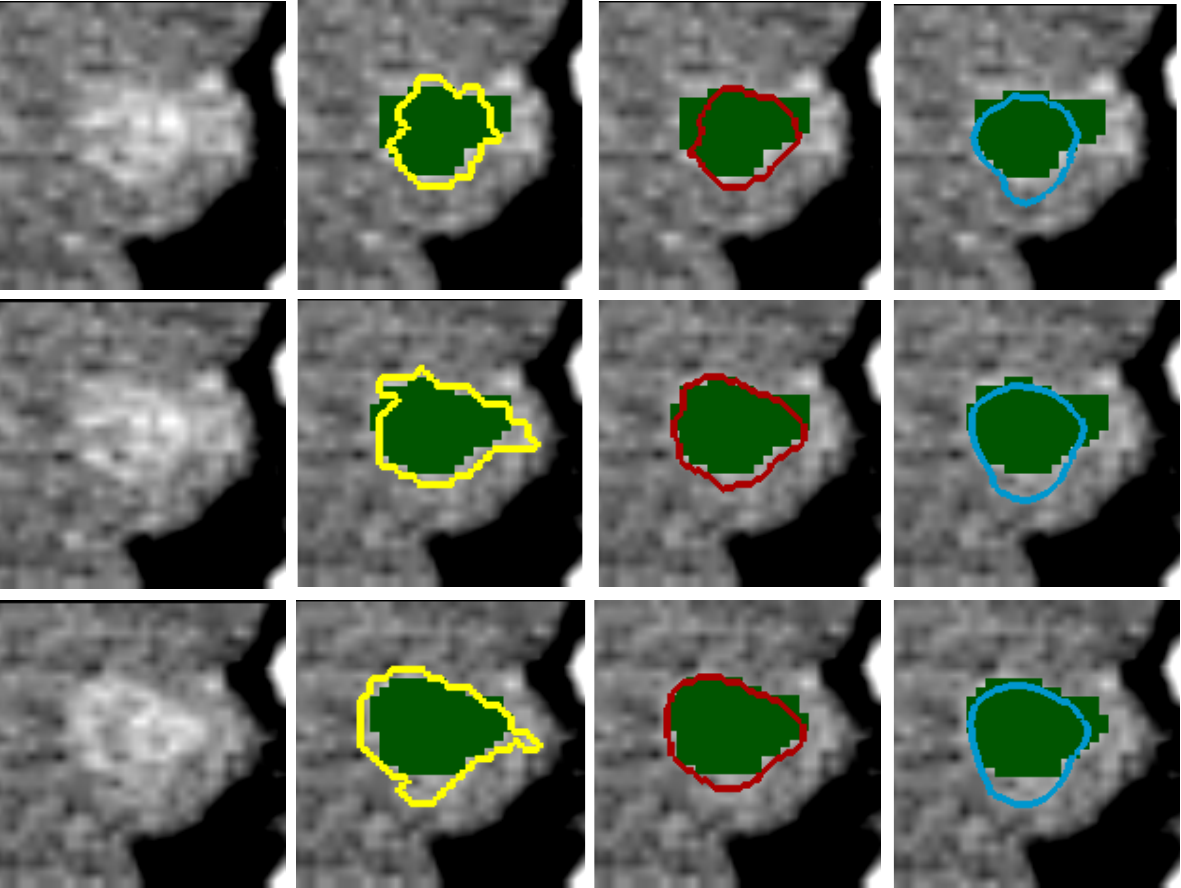}
\centering
\caption[example] 
{ \label{slices} 
Qualitative comparison of the three methods. The rows represent three adjacent slices from one tumor. Ground truth is marked as green regions. Yellow, red and blue contours are segmentations from the contextual UNet, deep LOGISMOS, and original LOGISMOS methods, respectively.}
\end{figure}

Segmentation results from contextual UNet after refinement, LOGISMOS and Deep LOGISMOS are presented in Fig.\ \ref{slices}. 
%------------------------------------------------ Figure ----------------------------------
%-------------------------------------------------------------------------------------------

%--------------------------------------------Table---------------------------------------
\begin{table}[t]
\centering
\caption{Comparison of segmentations from contextual UNet after refinement, LOGISMOS and Deep LOGISMOS.}
\label{tb:final}
\begin{tabular}{|c|c|c|}
\hline
Methods         & DSC (\%)            & RVD (\%)              \\ \hline
Contextual UNet & 81.6 $\pm$ 11.2 & 26.1 $\pm$ 20.9 \\ \hline
LOGISMOS        & 70.4 $\pm$ 27.7 & 35.4 $\pm$ 51.1 \\ \hline
\textbf{Deep LOGISMOS}   & \textbf{ 83.2 $\pm$ 7.8}  & \textbf{18.6 $\pm$ 17.4 } \\ \hline
\end{tabular}
\end{table}
%------------------------------------------------------------------------------------------

%------------------------------------------------ Figure ----------------------------------
\begin{figure}[t]
\includegraphics[width=6cm]{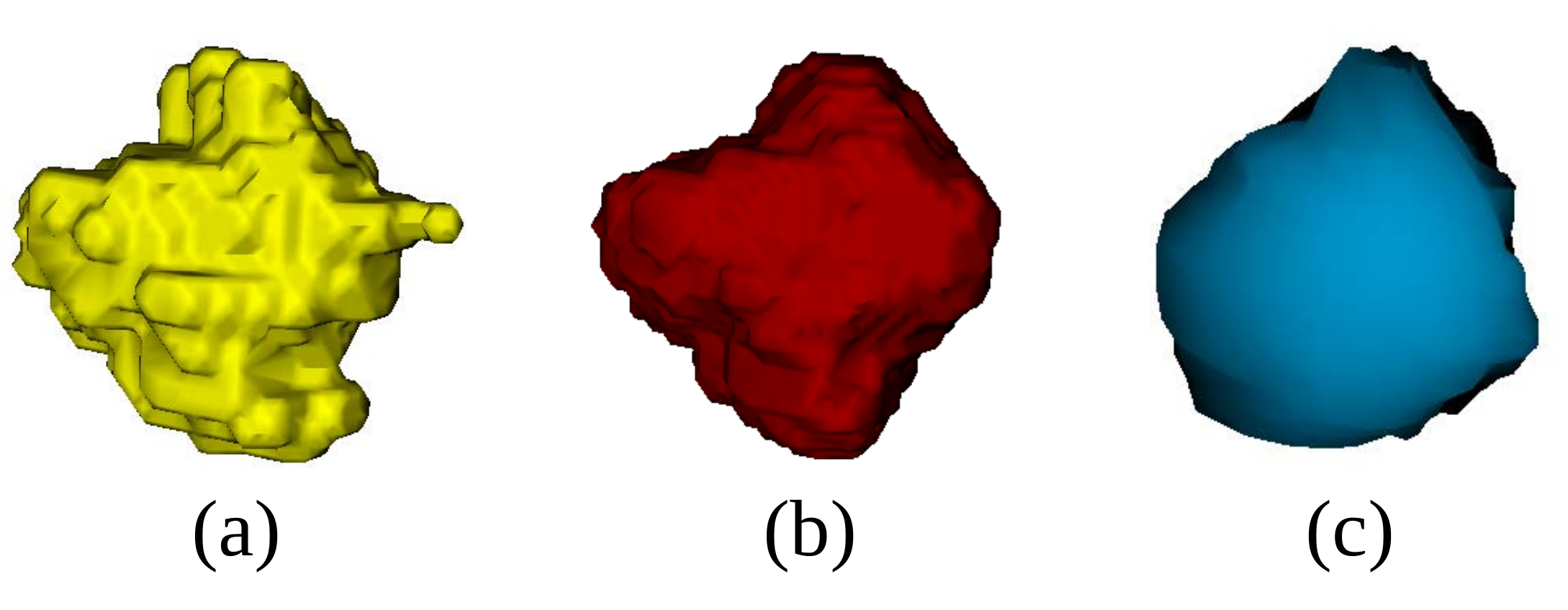}
\centering
\caption[example] 
{ \label{model} 
Tumor segmentations in the same 3D view; (a), (b) and (c) represent segmentations from contextual UNet, Deep LOGISMOS and original LOGISMOS respectively. The tumor is the same as in Fig.\ \ref{slices}.}
\end{figure} 
%-------------------------------------------------------------------------------------------
The initial segmentation for the original LOGISMOS method was a sphere centered on the ROI with the radius of 8 mm (a quarter of the ROI size), the costs were derived from inverted gradients along the graph columns. Deep LOGISMOS segmentation performance was significantly better than that of the contextual UNet with refinement and also of the original LOGISMOS when considering either of the  DSC and/or RVD metrics (Table \ref{tb:final}). Note that the LOGISMOS method failed to detect 3 tumors altogether since the tumors were too small. Excluding the 3 missing tumors, LOGISMOS method gave an average DSC of 78.7\%. The segmentation results demonstrated in a 3D view are shown in Fig.\ \ref{model}.
\section{Conclusion}
\label{sec:conclusion}
A hybrid fully convolutional network -- FCN combined with the graph-based LOGISMOS approach was reported. Its performance was evaluated in a 3D pancreatic  tumor segmentation task. 
Resulting from this study, we have demonstrated that 1)~context information from adjacent slices significantly improved the performance of a UNet, and that 2)~our novel Deep LOGISMOS method achieved significantly  better performance than the UNet and/or LOGISMOS methods alone. 

\bibliographystyle{IEEEbib}
\bibliography{strings,refs}

\end{document}